\documentclass[10pt,twocolumn,letterpaper]{article}

\usepackage{iccv}
\usepackage{times}
\usepackage{epsfig}
\usepackage{graphicx}
\usepackage{amsmath}
\usepackage{amssymb}

\graphicspath{{figures/}}
\usepackage{array}
\usepackage{makecell}
\usepackage{comment}
\usepackage{multirow}

\usepackage[breaklinks=true,bookmarks=false]{hyperref}

\iccvfinalcopy 

\def\iccvPaperID{****} 

\begin{document}

\title{Semantic-aware Image Deblurring}

\author{Fuhai Chen$^1$, Rongrong Ji$^1$\thanks{corresponding author}, Chengpeng Dai$^1$, Xiaoshuai Sun$^1$, Chia-Wen Lin$^2$, Jiayi Ji$^1$, \\ Baochang Zhang$^3$, Feiyue Huang$^4$, Liujuan Cao$^1$ \\
{\tt \small $^1$Xiamen University, $^2$National Tsing Hua University, $^3$Beihang University, $^4$Tencent Youtu Lab}
}

\maketitle

\begin{abstract}
Image deblurring has achieved exciting progress in recent years. However, traditional methods fail to deblur severely blurred images, where semantic contents appears ambiguously. In this paper, we conduct image deblurring guided by the semantic contents inferred from image captioning. Specially, we propose a novel Structured-Spatial Semantic Embedding model for image deblurring (termed S3E-Deblur), which introduces a novel Structured-Spatial Semantic tree model (S3-tree) to bridge two basic tasks in computer vision: image deblurring (ImD) and image captioning (ImC). In particular, S3-tree captures and represents the semantic contents in structured spatial features in ImC, and then embeds the spatial features of the tree nodes into GAN based ImD. Co-training on S3-tree, ImC, and ImD is conducted to optimize the overall model in a multi-task end-to-end manner. Extensive experiments on severely blurred MSCOCO and GoPro datasets demonstrate the significant superiority of S3E-Deblur compared to the state-of-the-arts on both ImD and ImC tasks.
\end{abstract}

\section{Introduction\label{sec:intro}}

\begin{figure}
\centering
\epsfig{file=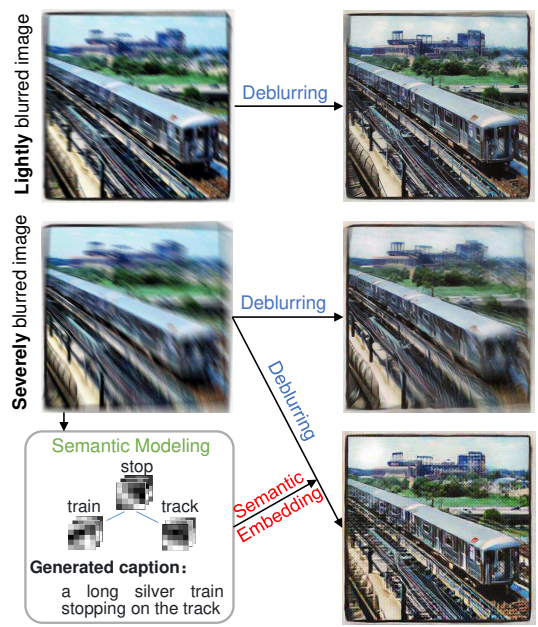, width=0.85\linewidth}
\caption{Image deblurring for images with light (Top) and severe (Middle \& Bottom) motion blur. The top and middle results are obtained by a state-of-the-art deblurring method \cite{kupyn2018deblurgan}, while the bottom one is generated by the proposed model with semantic (language) guidance.\label{fig:motivation}}
\end{figure}

Motion blur is an ubiquitous problem in photography, especially when using light-weight devices, such as mobile phones and on-board cameras. To remove the motion blur, many works have been proposed \cite{whyte2012non,ramakrishnan2017deep,gong2017motion,kupyn2018deblurgan,aittala2018burst}, which can successfully enhance images with light motion blur. An example is given in Fig. \ref{fig:motivation} (Top), where the sharp image is restored from the blurry one. In a typical setting, Generative Adversarial Networks (GANs) were recently adopted to estimate a adversarial loss and a feature consistency loss \cite{ledig2017photo} between the blurry and the sharp images \cite{ramakrishnan2017deep,kupyn2018deblurgan}.

\begin{figure*}
\centering
\vspace{-3mm}
\epsfig{file=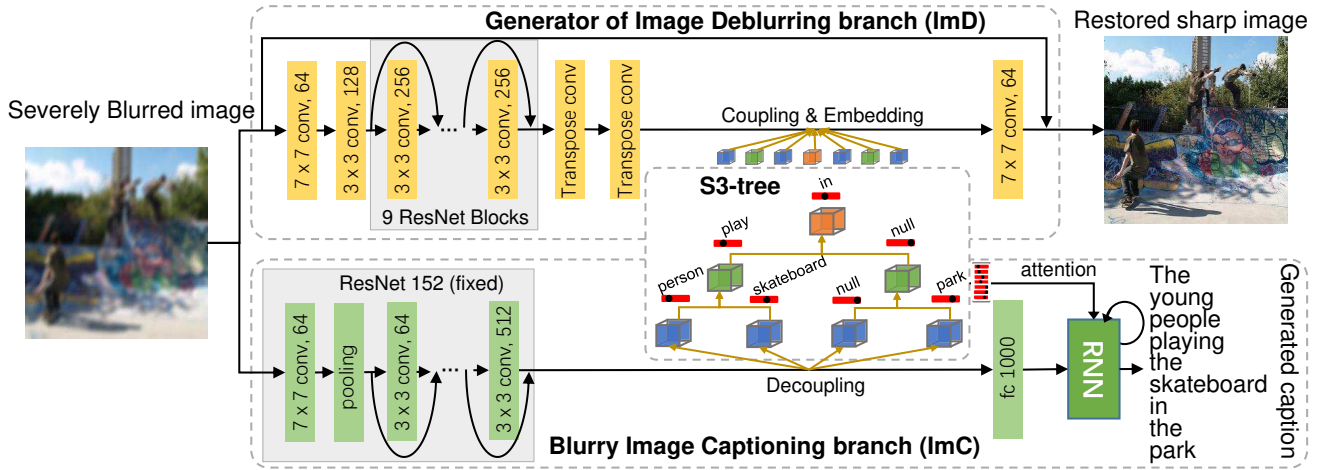, width=1.0\linewidth}
\caption{The overview of the proposed S3E-Deblur, which consists of S3-tree, image captioning branch (ImC), and image deblurring branch (ImD). Given a severely blurred image, its deep features are first extracted from CNN in ImD and ImC, respectively. Then the feature maps of ImC are decoupled into different semantic (entity/relation) spaces in S3-tree, where three operations are conducted in S3-tree, \emph{i.e.}, \emph{Convolutional decoupling} for the feature maps of entities (blue cubes), \emph{Convolutional combining} for the feature maps of relations (green and orange cubes), and \emph{Semantic classifying} for the probability distributions of entity/relation (red blocks), detailed in Sec. \ref{sec:sss-tree}. After that, the feature maps of tree nodes are coupled and embedded into the generator of GAN based ImD as well as attended into RNN of ImC. Finally, the classification loss of S3-tree, the reconstruction loss of ImC, the adversarial loss of ImD are jointly minimized to optimize the overall model in a multi-task end-to-end manner. \label{fig:framework}}
\vspace{-4mm}
\end{figure*}

Despite the exciting progress, the above methods can hardly deal with severely blurred images as shown in Fig. \ref{fig:motivation} (Middle). This is probably due to the ambiguity of the semantic contents derived from the rapid motion blur. We found that a sharper image can be restored given a right semantic guidance, \emph{e.g.}, \emph{track} and its related entity \emph{train} and their relation \emph{stop}. In fact, even given a severely blurred image with less information, human being can easily perceive the semantic contents as discussed in \cite{fei2007we}, and then reconstruct the scene in the brain \cite{horikawa2013neural,naselaris2009bayesian,shen2019deep} driven by the pattern of the evoked activities in the visual cortex \cite{huth2012continuous,stansbury2013natural}. On the other hand, inferring the semantic contents (entities and relations) is the core objective of many high-level semantic-related computer vision tasks like image captioning \cite{vinyals2015show,xu2015show,anderson2018bottom}. It is therefore a natural thought whether image captioning can be leveraged to guide image deblurring from a novel top-down manner.

In this paper, we aim to link image deblurring (ImD) with image captioning (ImC) to reinforce deblurring. To this end, we tackle two fundamental challenges, \emph{i.e.}, modeling semantics in ImC and embedding semantics to ImD as illustrated in Fig. \ref{fig:motivation} (Bottom). On the one hand, to overcome the arbitrary syntax of the caption \cite{chen2017structcap,wu2018interpretable}, we construct a structured semantic tree architecture, where the semantic contents (entities and relations) are automatically parsed given a severely blurred image. On the other hand, to align the semantic contents to the blurred image spatially, we design a spatial semantic representation for the tree nodes, where each entity/relation is represented in the form of feature maps, and the convolution is operated among the nodes in the tree structure.

In particular, we propose a \textbf{S}tructured-\textbf{S}patial \textbf{S}emantic \textbf{E}mbedding model for image deblurring, termed \emph{S3E-Deblur}, as illustrated in Fig. \ref{fig:framework}, where a \textbf{S}tructured-\textbf{S}patial \textbf{S}emantic tree (S3-tree) is constructed to bridge ImD and ImC. In particular, given a severely blurred image, its deep features are first extracted from convolutional neural network (CNN) in ImD and ImC, respectively. Then the feature maps of ImC are decoupled into different semantic (entity/relation) spaces by using the proposed S3-tree. After that, the feature maps of tree nodes are coupled and embedded into a convolutional layer in ImD. Simultaneously, the predicted probability distributions of semantic labels (entities/relations) are attended into the decoder (\emph{i.e.}, recurrent neural network (RNN)) for caption generation in ImC. Finally, S3-tree, ImC, and ImB are co-trained to optimize the overall model in a multi-task end-to-end manner.

The contributions of this paper are: 1) We are the first to import the semantic contents (from language) to guide image deblurring. 2) We propose a novel structured-spatial semantic tree model (S3-tree) to capture and represent the semantic contents to bridge image captioning (ImC) and image deblurring (ImD) branches. 3) We propose a multi-task end-to-end co-training scheme to restore the high-quality image and to generate its corresponding captions. (4) We release the first dataset of severely blurred images to facilitate the subsequent research.

\section{Related Work\label{sec:relatedWork}}
\paragraph{Image Deblurring.} The deblurring problems are divided into two types: blind and non-blind deblurring. Early works \cite{szeliski2010computer,schmidt2013discriminative,schuler2013machine,xu2014deep} mainly focused on non-blind deblurring, which assumed that the blur kernels are known. Recently, most deblurring works \cite{whyte2012non,schuler2016learning,chakrabarti2016neural,nah2017deep,ramakrishnan2017deep,gong2017motion,kupyn2018deblurgan,aittala2018burst} concerned about the more practical yet challenging case, \emph{i.e.}, blind deblurring, which typically adopted Convolutional Neural Networks (CNNs) with unknown blur kernels. In the aspect of blur types, the blurred images can be synthesized with uniform \cite{schuler2016learning,chakrabarti2016neural} or non-uniform \cite{nah2017deep,ramakrishnan2017deep,gong2017motion,kupyn2018deblurgan,aittala2018burst} blur kernels for training. The latter can deal with spatially-varying blurs, which has attracted extensive research attentions. For example, Gong \emph{et al.} \cite{gong2017motion} adopted a fully convolutional network (FCN) \cite{long2015fully} to estimate the motion flow following with a non-blind deconvolution. However, the prior map of motion flow map should be provided during training. To overcome this defect, Aittala \emph{et al.} \cite{aittala2018burst} proposed a U-Net \cite{ronneberger2015u} based end-to-end encoder-decoder model for both video and single-image deblurring. Kupyn \emph{et al.} \cite{kupyn2018deblurgan} proposed a GAN based image deblurring with a multi-component loss function \cite{ledig2017photo}, which achieved the state-of-the-art result on the non-uniform blind deblurring. However, all the above works focused on deblurring lightly blurred images, which cannot well handle severely blurred images with ambiguous semantic contents.

\paragraph{Image Captioning.} Image captioning has recently attracted extensive research attention. Most existing image captioning methods were inspired by the encoder-decoder framework in machine translation \cite{bahdanau2014neural,sutskever2014sequence}. From this perspective, image captioning is analogous to translating images to texts. As the mainstream of general image captioning, Vinyals \emph{et al.} \cite{vinyals2015show} and Karpathy \emph{et al.} \cite{karpathy2015deep} proposed a CNN-RNN architecture, where the visual features are extracted by a CNN, and then fed into RNNs to output word sequences as captions. Based on the CNN-RNN architecture, Xu \emph{et al.} \cite{xu2015show} and You \emph{et al.} \cite{you2016image} proposed to use an attention module in captioning based on the spatial features and the semantic concepts, respectively. To capture the semantic entities and their relations, Chen \emph{et al.} \cite{chen2017structcap} proposed a visual parsing tree model to embed and attend the structured semantic content into captioning. To advance the attention model, a bottom-up and top-down attention mechanism was designed in \cite{anderson2018bottom}, which enabled attention to be calculated at the object/region level. It achieved the state-of-the-art results.

\section{The Proposed Method\label{sec:method}}

\begin{figure}
\centering
\vspace{-5mm}
\epsfig{file=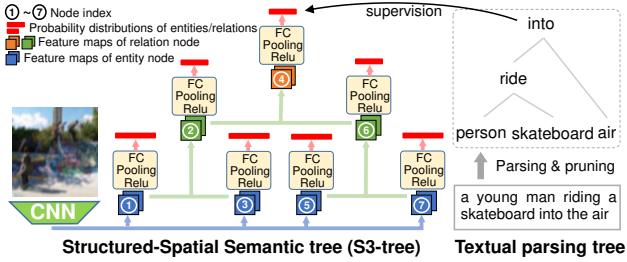, width=1.0\linewidth}
\caption{The architecture of S3-tree. The fundamental operations include \emph{Convolutional decoupling} (wathet arrow), \emph{Convolutional combining} (aqua arrow), and \emph{Semantic classifying} (light red arrow). Nodes with indexes \textcircled{\footnotesize{1}}$\thicksim$\textcircled{\footnotesize{7}}: subject 1, sub-relation 1, object 1, root-relation, subject 2, sub-relation 2, and object 2. \label{fig:SSS-tree}}
\vspace{-3mm}
\end{figure}

The goal of the proposed S3E-Deblur is to deblur a severely blurred image. The framework is illustrated in Fig. \ref{fig:framework}, which consists of image captioning branch (ImC), image deblurring (ImD), and the S3-tree. Specially, S3-tree is first modeled (Sec. \ref{sec:sss-tree}) to capture and represent the semantic contents in ImC. Then the structured spatial semantic features are embedded into ImD (Sec. \ref{sec:ssse-deblur}). Finally, S3-tree, ImD, and ImC are jointly co-trained in a multi-task end-to-end manner (Sec. \ref{co-training}).

\subsection{S3-tree Modeling\label{sec:sss-tree}}

In the blurred image captioning, we aim to learn the parameters of S3-tree model (defined as $\mathbb{T}$) given a pair of a blurry image $I^B$ and a caption $S$ describing this image, where the tree model can be used to automatically infer the structured contents and provide the structured spatial features during inference. The architecture of S3-tree is illustrated in Fig. \ref{fig:SSS-tree}, which consists of three operations: \emph{Convolutional decoupling}, \emph{Convolutional combining}, and \emph{Semantic Classifying}. We itemize them as follows:

\begin{figure}
\centering
\vspace{-5mm}
\epsfig{file=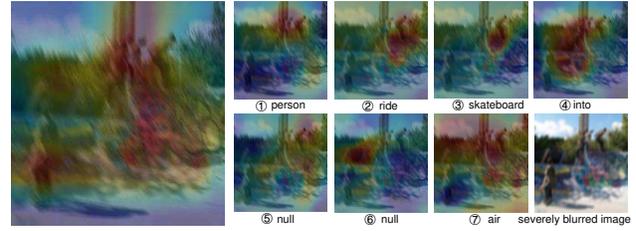, width=1.0\linewidth}
\caption{Heatmaps of the feature maps in the last convolution layer of ResNet (Left) and the nodes of S3-tree (Right). Node indexes corresponds to Fig. \ref{fig:SSS-tree}.\label{fig:decoupling}}
\vspace{-4mm}
\end{figure}

\paragraph{Convolutional Decoupling.} The visual feature map in a tensor form ${\rm\mathbf{V}}\in \mathcal{R}^{w\times h\times c}$ of $I^B$ is first extracted from the last convolutional layer of CNN (ResNet-152) \cite{simonyan2014very}, where $w$, $h$, and $c$ denote the width, height, and channel of the tensor, respectively. Then we convolute ${\rm\mathbf{V}}$ into different semantic spaces, \emph{i.e.}, subjects, objects, and relations to decouple the semantic content, which can be formulated as:

\vspace{-5mm}
\begin{align}
&\quad\quad\quad {\rm\mathbf{H}}_{[\cdot,\cdot,q]}^{j} = \sum_{p=0}^{c}\sigma({\rm\mathbf{K}}_{p,q}^{E}\circledast{\rm\mathbf{V}}_{[\cdot,\cdot,p]}), \\[-1mm]
&\resizebox{.84\linewidth}{!}{$s.t.(j:E) \in \{(1:Subj1),(3:Obj1),(5:Subj2),(7:Obj2)\},$}\nonumber
\end{align}
\noindent
where $j$ is the index of the tree node as shown in Fig. \ref{fig:SSS-tree}, and ${\rm\mathbf{H}}^{j}\in \mathcal{R}^{w'\times h'\times c'}$ ($w'$, $h'$, and $c'$ denote the width, the height, and the channel, respectively) is a feature map tensor of the $j$-th node. $E$ represents one of four semantic entities, \emph{i.e.}, subject 1 (Subj1), object 1 (Obj1), subject 2 (Subj2), and object 2 (Obj2) as set up in \cite{chen2017structcap}. $\sigma$ is an element-wise nonlinear function upon the convolution operation $\circledast$ and the convolution kernel set ${\rm\mathbf{K}}^{E}$ for \emph{Convolutional decoupling} in the $j$-th node ($j=1,3,5,7$). There are $c\cdot c'$ kernels with size $3\times 3$ in ${\rm\mathbf{K}}^{E}$.

\paragraph{Convolutional Combining.} The feature map tensors of two child nodes, \emph{i.e.}, ${\rm\mathbf{H}}^{j'}$ and ${\rm\mathbf{H}}^{j''}$ of the $j'$-th and the $j''$-th nodes, are convoluted into a combined feature map tensor ${\rm\mathbf{H}}^j$ of their parent (the $j$-th) node, formulated as follows:

\vspace{-5mm}
\begin{align}
&\quad\quad\quad {\rm\mathbf{H}}_{[\cdot,\cdot,q]}^j = \sum_{p=0}^{2\cdot c'}\sigma\big({\rm\mathbf{K}}_{p,q}^{R}\circledast[{\rm\mathbf{H}}^{j'}; {\rm\mathbf{H}}^{j''}]_{[\cdot,\cdot,p]}\big), \\[-1mm]
&\resizebox{1\linewidth}{!}{$s.t.(j:R\triangleleft j',j'') \in \{(2:sRel1\triangleleft 1,3),(6:rRel\triangleleft 5,7),(4:sRel2\triangleleft 6,7)\},$} \nonumber
\end{align}
\noindent
where $R$ represents one of the three semantic relation items, \emph{i.e.}, sub-relation 1 (sRel1), root-relation (rRel), and sub-relation 2 (sRel2) as set up in \cite{chen2017structcap}. $\triangleleft$ denotes the combination, \emph{e.g.}, the 1-st and the 3-rd nodes are combined into the 2-nd node in $(2:sRel1\triangleleft 1,3)$. ${\rm\mathbf{H}}^j$ has the same size of ${\rm\mathbf{H}}^{j'}$ and ${\rm\mathbf{H}}^{j''}$. $[\cdot;\cdot]$ is the feature map-wise concatenation operation. There are $2\cdot c'\cdot c'$ kernels with size $3\times 3$ in ${\rm\mathbf{K}}^{R}$.

\paragraph{Semantic Classifying.} Upon the pre-processed entity/relation vocabulary, the feature map tensor ${\rm\mathbf{H}}$ of each node is mapped into the entity/relation category space, where ${\rm\mathbf{H}}$ is firstly transformed into the feature vector ${\rm\mathbf{h}}$ via a Relu based nonlinear function \cite{nair2010rectified}, an average-pooling layer and a fully-connected layer. We denote the parameter set of the aforementioned three layers by ${\rm\mathbf{W}}$. Thereout, we denote the parameter set of S3-tree model $\mathbb{T}$ by $\psi = \{{\rm\mathbf{K}}^{E},{\rm\mathbf{K}}^{R},{\rm\mathbf{W}}|E \in \{Subj1,Obj1,Subj2,Obj2\}, R \in \{sRel1,rRel,sRel2\}\}$. To optimize $\mathbb{T}_{\psi}$, the predicted entities/relations are supervised by the corresponding labels in the textual parsing tree\footnote{Textual parsing and pruning preprocesses are conducted on the caption following \cite{chen2017structcap} for the fixed tree structure.}. Specially, ${\rm\mathbf{h}}$ of each node is used to compute the cross entropy loss with the entity/relation category label, which can be formulated as:

\vspace{-3mm}
\begin{equation}
\mathcal{L}_{\mathbb{T}} = -\sum_{j=0}^M\log {\rm P}(y^j|{\rm\mathbf{V}};\psi),
\label{equ:tree_loss}
\end{equation}
\noindent
where $y^j$ denotes the entity/relation category in the $j$-th node. $M=7$ denotes the number of nodes. The predicted label vectors can be represented as $\{{\rm\widehat{\mathbf{y}}}_j\}_{j=1}^M$. For learning effective representations, we pre-train S3-tree based upon the above loss before the overall model training. After that, we visualize the feature maps from ResNet and nodes in Fig. \ref{fig:decoupling}, where we find the former is well parsed into the semantic spaces.

\subsection{S3-tree Embedding in ImD\label{sec:ssse-deblur}}

\begin{figure*}
\centering
\vspace{-4mm}
\epsfig{file=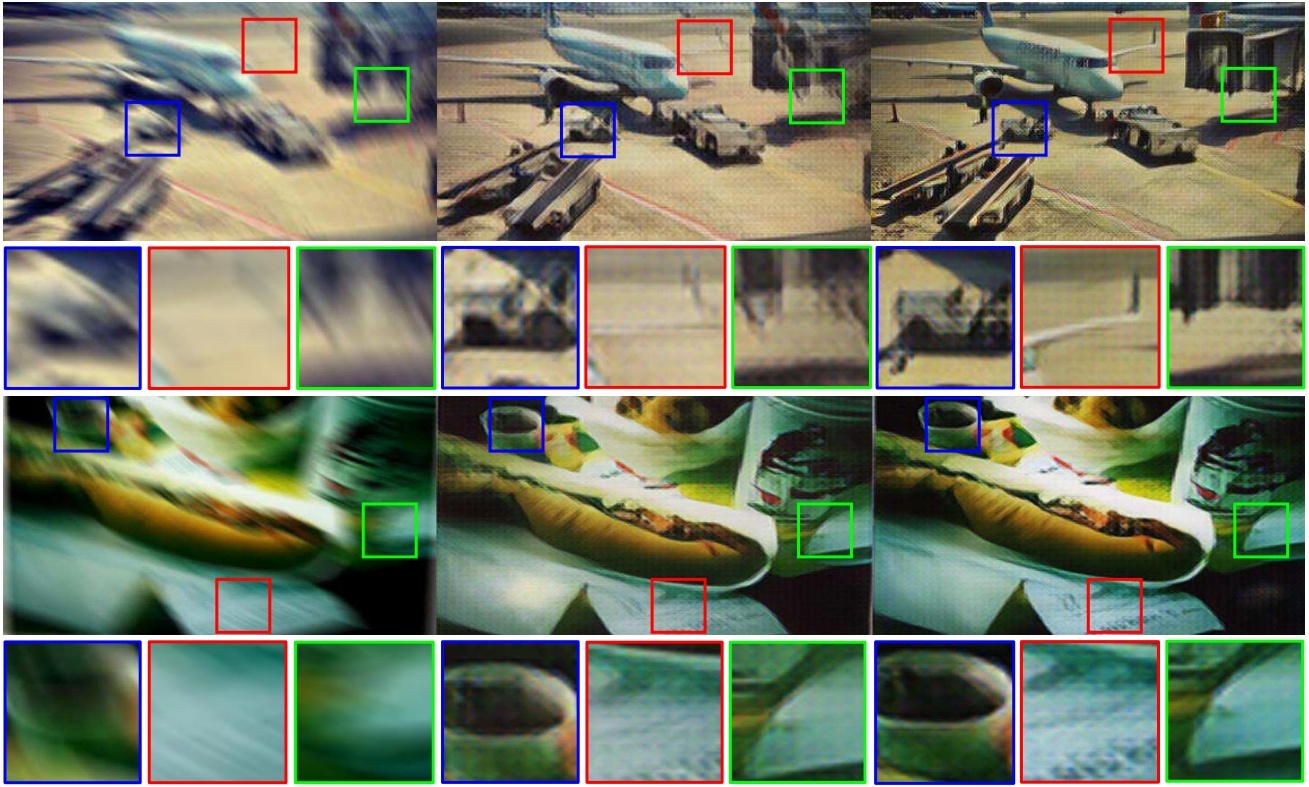, width=1.0\linewidth}
\caption{Deblurring results on Sev-BlurData dataset. From left to right: The severely blurred images, the restored images via DeblurGAN \cite{kupyn2018deblurgan}, and the restored images via the proposed S3E-Deblur. \label{fig:visSID_cocoBased}}
\vspace{-3mm}
\end{figure*}

Given a blurry image $I^B$, we first obtain the set of semantic feature map tensors $\{{\rm\mathbf{H}}^j\}_{j=1}^{M}$ via S3-tree model $\mathbb{T}_{\psi}({\rm\mathbf{V}})$, where ${\rm\mathbf{V}}$ is the fixed feature map tensor from CNN as aforementioned in Sec. \ref{sec:sss-tree}. Then we embed the semantic feature map tensors into a GAN based deblurring model. The loss function is formulated as:

\vspace{-2mm}
\begin{equation}
\mathcal{L}_{\text{\tiny ImD}-\mathbb{T}} = \mathcal{L}_{\text{\tiny GAN}-\mathbb{T}} + \lambda \cdot \mathcal{L}_{\text{\tiny X}-\mathbb{T}},
\end{equation}
\noindent
where $\lambda$ is a trade-off coefficient (set as 100). $\mathcal{L}_{\text{\tiny GAN}-\mathbb{T}}$ is a S3-tree guided adversarial loss while $\mathcal{L}_{\text{\tiny X}-\mathbb{T}}$ is a S3-tree guided content loss. We itemize them as below.

\paragraph{S3-tree Guided Adversarial Loss.} Recently, many works apply conditional GANs for image-to-image translation, which aim to fool the discriminator network \cite{li2016precomputed,zhu2017unpaired,ledig2017photo,bousmalis2017unsupervised}. We use WGAN-GP \cite{gulrajani2017improved} as the critic function in our deblurring model, which is shown to be robust to the choice of the generator \cite{arjovsky2017wasserstein}. The loss can is calculated as follows:

\vspace{-2mm}
\begin{equation}
\mathcal{L}_{\text{\tiny GAN}-\mathbb{T}} = -D_{\theta}\Big(G_{\phi}\big(I,\mathbb{T}_{\psi}({\rm\mathbf{V}})\big)\Big),
\end{equation}
\noindent
where the generative loss is defined based on the discriminator $D_{\theta}$, the generator $G_{\phi}$, and the S3-tree $\mathbb{T}_{\psi}$, with their parameters $\theta$, $\phi$, and $\psi$, respectively. It's noted that the mismatch penalization between the input and the output, as in \cite{zhu2017unpaired}, is removed due to the inherent inconsistency between the severely blurred image and the restored sharp image. Similar to \cite{johnson2016perceptual}, we adopt two strided convolution blocks with stride $\frac{1}{2}$, nine residual blocks (ResBlocks) \cite{he2016deep} and two transposed convolution blocks \cite{radford2015unsupervised} in $G_{\phi}$, as shown in Fig. \ref{fig:framework}. Additionally, we add the skip connections \cite{ledig2017photo,kupyn2018deblurgan} among the blocks of $G_{\phi}$ for better feature fusion. In the last convolution block, we couple the feature maps $\{{\rm\mathbf{H}}^j\}_{j=1}^{M}$ of S3-tree $\mathbb{T}_{\psi}$ and integrate them with the input feature maps of the convolution layer ($L$-th layer). We compute the $q$-th input feature map as:

\vspace{-3mm}
\begin{equation}
{\rm\mathbf{H}}_{[\cdot,\cdot,q]}^L = [{\rm\mathbf{H}}_{[\cdot,\cdot,q]}^{L};\sum_{p=0}^{M\cdot c'}\sigma({\rm\mathbf{K}}_{p,q}^{L}\circledast[\{{\rm\mathbf{H}}^j\}_{j=1}^{M}]_{[\cdot,\cdot,p]})],
\end{equation}
\noindent
where $c'$, $\circledast$, and $\sigma$ denote the feature map channel of each node, the convolution operation, and an element-wise nonlinear function, respectively, as aforementioned in Sec. \ref{sec:sss-tree}. We define $[\{\cdot\}]$ as a concatenation operation of the elements in a set. ${\rm\mathbf{K}}^{L}$ is a convolution kernel, which contains $c'\cdot c''$ ($c''$ is the channel of the $L$-layer feature maps) kernels with size $3\times 3$.

\paragraph{S3-tree Guided Content Loss.} Inspired by the pixel-wise mean
squared error (MSE) \cite{dong2016image,shi2016real}, we adopt an advanced Perceptual loss \cite{johnson2016perceptual,ledig2017photo} as the basis of our content loss, which encourages the restored image guided by S3-tree $\mathbb{T}_{\psi}$ to have similar perceptual feature to the sharp image. We formulate our content loss as:

\vspace{-3mm}
\begin{equation}
\mathcal{L}_{\text{\tiny X}-\mathbb{T}} = \frac{1}{w_o\cdot h_o}\Big\|F_l(I^S)-F_l\Big(G_{\phi}\big(I^B,\mathbb{T}_{\psi}({\rm\mathbf{V}})\big)\Big)\Big\|_2^2,
\end{equation}
\noindent
where $F_l$ is a CNN based network, which outputs the feature map in the $l$-th layer as the perceptual feature representation. We adopt the part of VGG-19 network \cite{simonyan2014very} (pretrained on ImageNet \cite{Deng2009ImageNet}) before the $3$-rd convolution layer as $F_l$. $w_o$ and $h_o$ denote the width and the height of the output feature map, respectively. The perceptual loss can restore the general contents \cite{isola2017image,ledig2017photo} while adversarial loss can restore the texture details.

\subsection{Co-training\label{co-training}}

For the overall semantic understanding, we integrate the S3-tree $\mathbb{T}$ into the mainstream backbone of image captioning, \emph{i.e.}, CNN-RNN \cite{vinyals2015show}. Specially, we first apply the attention mechanism \cite{you2016image,xu2015show} to attend the predicted label vectors $\{{\rm\widehat{\mathbf{y}}}_j\}_{j=1}^M$ of $\mathbb{T}$ in Eq. \ref{equ:tree_loss} into each hidden state of RNN (To be exact, LSTM \cite{hochreiter1997long,vinyals2015show}). Then we adopt the commonly-used negative log likelihood as the reconstruction loss to train the blurred image captioning model based on $\mathbb{T}$:

\vspace{-4mm}
\begin{equation}
\mathcal{L}_{\text{\tiny ImC}-\mathbb{T}}=-\sum_{t=0}^{T}\log p\big(S_t|{\rm\mathbf{h}}_{\rm\mathbf{V}},\{{\rm\widehat{\mathbf{y}}}_j\}_{j=1}^M,S_{0:t-1}\big),
\label{equ:IC_loss}
\end{equation}
\noindent
where ${\rm\mathbf{h}}_{\rm\mathbf{V}}$ is the feature vector from ${\rm\mathbf{V}}$ via a pooling and a fully-connected layers. $S$ denotes the caption with $T$ words. The $t$-th word $S_t$ is generated on the previous words $S_{0:t-1}$ and the feature vector set of $\mathbb{T}_{\psi}$. Finally, we have the total loss for severely image deblurring, blurred image captioning, and S3-tree as:

\vspace{-2mm}
\begin{equation}
\mathcal{L}_{\text{\tiny Total}} = \mathcal{L}_{\text{\tiny ImD}-\mathbb{T}} + \lambda_{\text{\tiny ImC}} \mathcal{L}_{\text{\tiny ImC}-\mathbb{T}} + \lambda_{\mathbb{T}} \mathcal{L}_{\mathbb{T}}.
\end{equation}
\noindent
where $\lambda_{\text{\tiny ImC}}$ and $\lambda_{\mathbb{T}}$ are trade-off coefficients (set as $5e^{-3}$ and $10$, respectively). For optimization, we follow the approach of \cite{arjovsky2017wasserstein} and perform 5 gradient descent steps on $D_{\theta}$, then one step on $\{G_{\phi},\mathbb{T}_{\psi}\}$ by adopting Adam \cite{kingma2014adam} as a solver. The learning rate is initially set to $10^{-4}$ for both generator and critic. After the first 150 epochs, we linearly decay the rate to zero over the next 150 epochs. The method is implemented in PyTorch\footnote{\url{http://pytorch.org}}, which takes roughly 58 training hours on NVIDIA GTX 1080 Ti GPU. During online inference, we follow the idea of \cite{zhu2017unpaired} and apply both dropout and instance normalization.

\section{Experiments\label{sec:exps}}

\begin{table*}\small
\centering
\vspace{-3mm}
\caption{Performance comparisons of ImD on LessSev-BlurData and Sev-BlurData.}
\begin{tabular}{l|c|c|c|c|c|c}
\hline
& \multicolumn{3}{c|}{LessSev-BlurData} & \multicolumn{3}{c}{Sev-BlurData} \\
\hline
 & U-Deconv \cite{aittala2018burst} & DeblurGAN \cite{kupyn2018deblurgan} & S3E-Deblur (ImD) & U-Deconv \cite{aittala2018burst} & DeblurGAN \cite{kupyn2018deblurgan} & S3E-Deblur (ImD) \\
\hline
PSNR &  21.88 & 22.36 & 23.35  &  21.71 & 21.14  &  22.87 \\
SSIM & 0.680  & 0.705  & 0.712  & 0.675  & 0.693  & 0.702  \\
\hline
\end{tabular}
\label{tab:comparison_sota_SID}
\end{table*}

\begin{figure*}
\centering
\epsfig{file=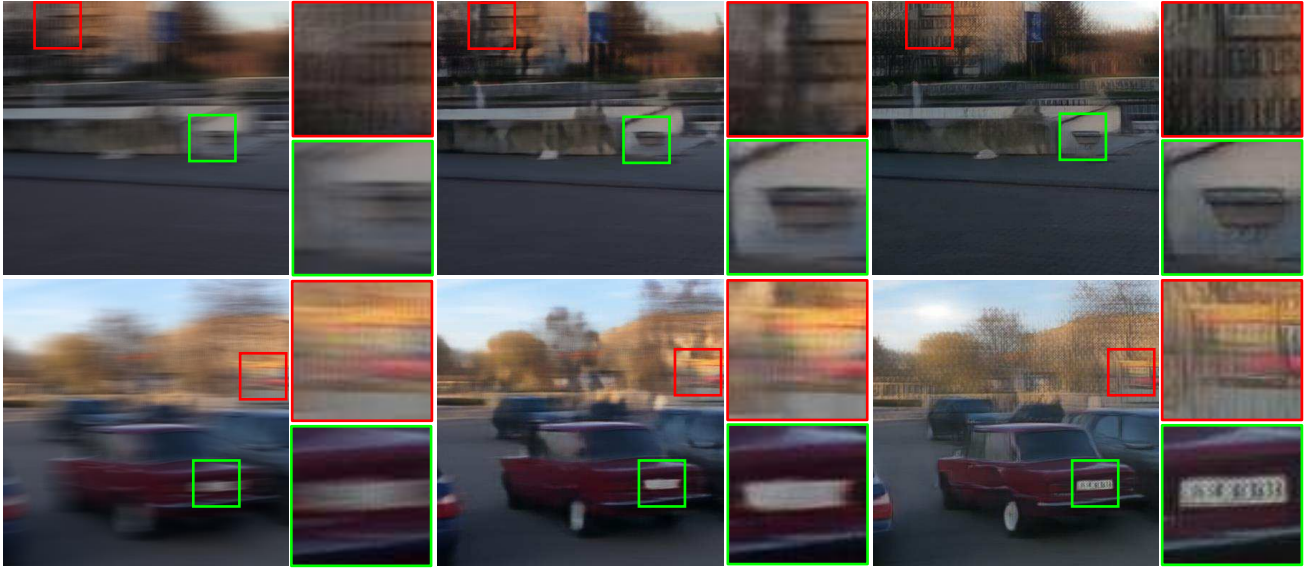, width=1.0\linewidth}
\caption{Deblurring results on the severely blurred GoPro dataset. From left to right: The severely blurred images, the restored images via DeblurGAN \cite{kupyn2018deblurgan}, and the restored images via the proposed S3E-Deblur. \label{fig:visSID_otherBased}}
\end{figure*}

In this section, we first introduce the datasets and the experimental setting in Sec. \ref{sec:dataset_setting}. Then we compare both the quantitative and qualitative results of our proposed model to the competing methods on severely image deblurring in Sec. \ref{sec:eval_SID}. Additionally, we evaluate the semantic understanding ability of the proposed S3-tree on blurred image captioning. Finally, we conduct ablation study for the proposed S3-tree in Sec. \ref{sec:analysis_SSS-tree}.

\subsection{Dataset and Setting\label{sec:dataset_setting}}

\paragraph{Datasets.} We first introduce two new benchmark datasets containing severely blurred images for deblurring and captioning, which are the first of their kind. Each of them contains 13K blurry images, 13K sharp images and 65K captions (5 captions per blurry/sharp image) derived from the public split\footnote{\url{https://github.com/karpathy/neuraltalk}} of the MSCOCO dataset \cite{chen2015microsoft}, where each blurry image is deblurred via simulating motion flow\footnote{\url{https://github.com/donggong1/motion-flow-syn}} in \cite{gong2017motion}. The images in these two datasets are blurred with two random Gaussian noise ranges, $[0.2,0.5]$ and $[0.5,1]$, respectively, which are termed LessSev-BlurData and Sev-BlurData\footnote{Two datasets will be released publicly.}. Additionally, we conduct model testing on another benchmark, \emph{i.e.}, GoPro dataset \cite{nah2017deep}, where the images are severely blurred in the above way. The dataset consists of 2K pairs of the blurry and sharp images derived from various scenes.

\paragraph{Preprocessing.} We parse the captions by using the Stanford Parser \cite{socher2011parsing} and then prune the textual parsing results by using the pos-tag tool and the lemmatizer tool in NTLK \cite{loper2002nltk}. During pruning, the dynamic textual parsing trees are converted to a fixed, three-layer, binary tree as designed in \cite{chen2017structcap}. Only the words with high frequency are left to form the vocabularies. Nouns are regarded as entities and used as the leaf nodes in the S3-tree, while others (verbs, coverbs, prepositions, and conjunctions) are taken as relations for the non-leaf nodes. To reduce the sparsity of the vocabularies, we combine the verb and the symbol ``\_P'' for the coverb. ``null'' is added to the entity vocabulary for the absence of word and the lowly frequent words. The sizes of the entity and the relation vocabularies are finally 839 and 247, respectively, where similar words are merged into one word by employing WordNet \cite{miller1995wordnet}.

\begin{table}\small
\centering
\caption{Performance comparisons of ImD via testing models on severely blurred GoPro dataset. ``w coTrain'' denotes co-training with S2-tree in ImC, and vice versa for ``w/o coTrain''.}
\begin{tabular}{l|c|c|c|c}
\hline
& U-Deconv & DeblurGAN & \multicolumn{2}{c}{S3E-Deblur (ImD)} \\ \cline{4-5}
& \cite{aittala2018burst} & \cite{kupyn2018deblurgan} & w/o coTrain &  w coTrain \\
\hline
PSNR &  20.85 &  20.21 &  22.12  &  23.60 \\
SSIM &  0.628 &  0.631 &  0.731  &  0.744 \\
\hline
\end{tabular}
\label{tab:comparison_sota_SID_transfer}
\end{table}

\begin{table*}\small
\centering
\vspace{-3mm}
\caption{Performance comparisons of ImC on the sharp data, LessSev-BlurData and Sev-BlurData derived from MSCOCO dataset.}
\begin{tabular}{c|c|c|c|c|c|c|c|c|c}
\hline
Dataset & Method & Bleu-1 & Bleu-2 & Bleu-3 & Bleu-4 & CIDEr & Meteor & Rouge-L & Spice \\
\hline
\multirow{5}*{Sharp data}
& NIC \cite{vinyals2015show}        & 0.725 & 0.557 & 0.422 & 0.320 & 0.991 & 0.253 & 0.535 & 0.184 \\
& Toronto \cite{xu2015show}         & 0.747 & 0.584 & 0.443 & 0.333 & 1.028 & 0.256 & 0.547 & 0.190 \\
& VP-tree \cite{chen2017structcap}  & 0.743 & 0.580 & 0.442 & 0.336 & 1.033 & 0.261 & 0.549 &  -  \\
& TopDown \cite{anderson2018bottom} & 0.752 & 0.588 & 0.449 & 0.347 & 1.057 & \textbf{0.269} & 0.551 & \textbf{0.200} \\
& S3E-Deblur (ImC)                   & \textbf{0.768}& \textbf{0.610} & \textbf{0.471} & \textbf{0.359} & \textbf{1.106} & 0.267 & \textbf{0.564} & 0.195 \\
\hline
\multirow{5}*{LessSev-BlurData}
& NIC \cite{vinyals2015show}        & 0.692 & 0.521 & 0.388 & 0.291 & 0.866 & 0.233 & 0.510 & 0.163 \\
& Toronto \cite{xu2015show}         & 0.706 & 0.536 & 0.398 & 0.295 & 0.888 & 0.234 & 0.517 & 0.167 \\
& VP-tree \cite{chen2017structcap}  & 0.695 & 0.527 & 0.392 & 0.291 & 0.899 & 0.244 & 0.520 & 0.176 \\
& TopDown \cite{anderson2018bottom} & 0.689 & 0.521 & 0.389 & 0.292 & 0.917 & 0.249 & 0.522 & 0.179 \\
& S3E-Deblur (ImC)                   & \textbf{0.739} & \textbf{0.582} & \textbf{0.449} & \textbf{0.345} & \textbf{1.079} & \textbf{0.265} & \textbf{0.563} & \textbf{0.191} \\
\hline
\multirow{5}*{Sev-BlurData}
& NIC \cite{vinyals2015show}        & 0.639 & 0.461 & 0.335 & 0.248 & 0.696 & 0.209 & 0.473 & 0.138 \\
& Toronto \cite{xu2015show}         & 0.661 & 0.485 & 0.352 & 0.257 & 0.729 & 0.211 & 0.486 & 0.141 \\
& VP-tree \cite{chen2017structcap}  & 0.646 & 0.468 & 0.335 & 0.243 & 0.726 & 0.219 & 0.485 & 0.151 \\
& TopDown \cite{anderson2018bottom} & 0.647 & 0.470 & 0.340 & 0.249 & 0.753 & 0.223 & 0.490 & 0.153 \\
& S3E-Deblur (ImC)                   & \textbf{0.703} & \textbf{0.536} & \textbf{0.406} & \textbf{0.309} & \textbf{0.938} & \textbf{0.254} & \textbf{0.534} & \textbf{0.178} \\
\hline
\end{tabular}
\label{tab:comparison_sota_BIC}
\end{table*}

\begin{figure*}
\centering
\vspace{-2mm}
\epsfig{file=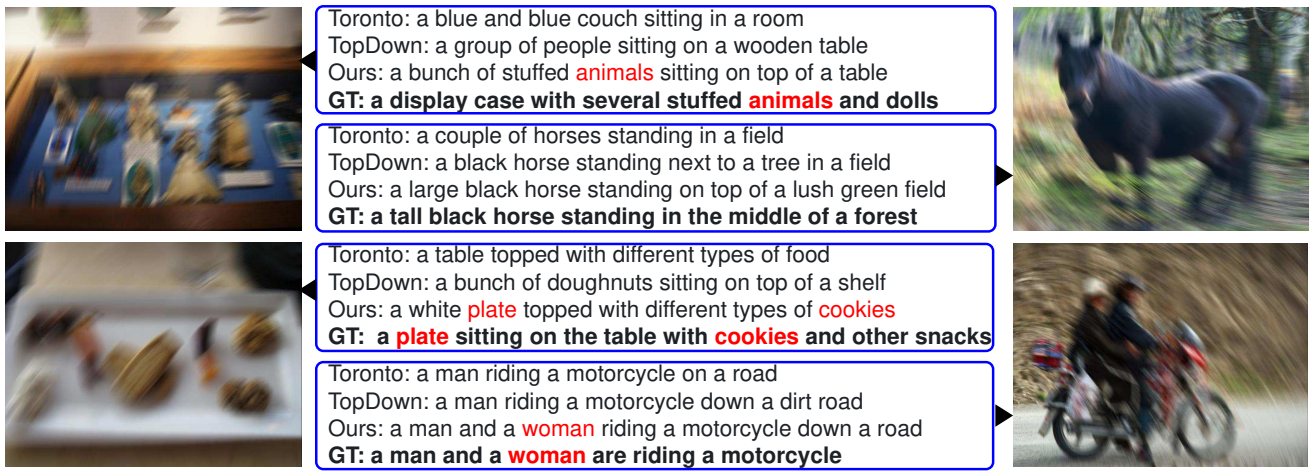, width=1.0\linewidth}
\caption{Captioning results of Toronto \cite{xu2015show}, TopDown \cite{anderson2018bottom}, and our S3E-Deblur (ImC), respectively, on Sev-BlurData. The ground truth captions (GT) are in bold font. The unique unique semantic entities recognized by S3E-Deblur (ImC) are marked in red. \label{fig:VisBIC}}
\vspace{-3mm}
\end{figure*}

\begin{figure*}
\centering
\vspace{-3mm}
\epsfig{file=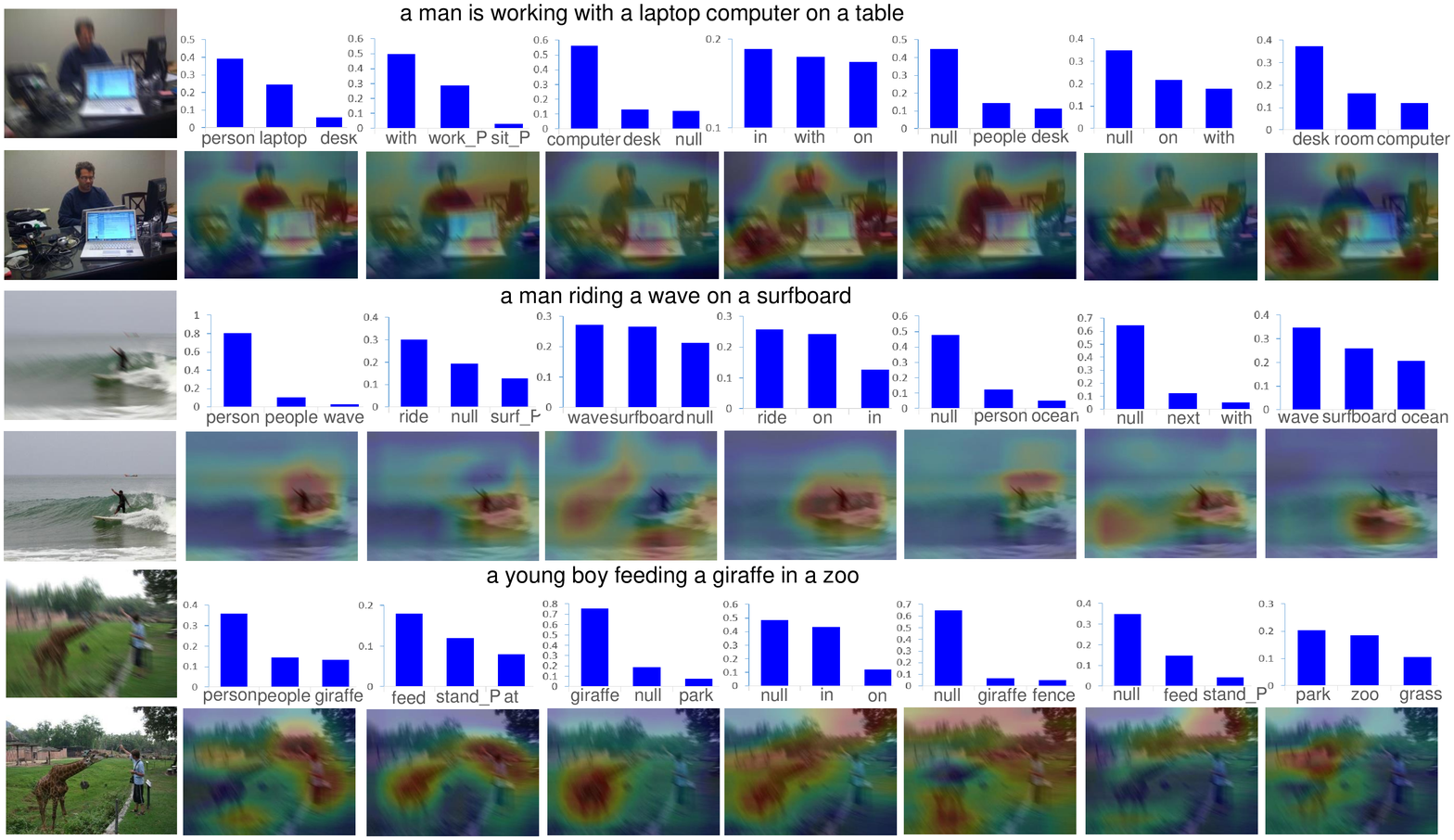, width=1.0\linewidth}
\caption{Output visualization of the S3-tree. The severely blurred and the shape images are in the first column. The other seven columns are the outputs of S3-tree nodes corresponding to the node indexes in Fig. \ref{fig:SSS-tree}. Each column presents the top-3 probability values of the entity/relation categories and the heatmap on the feature maps of each node. Generated captions are presented in the top of rows. \label{fig:visSSS-tree}}
\vspace{-3mm}
\end{figure*}

\paragraph{Metrics.} To evaluate the restored images, we adopt peak signal-to-noise ratio (PSNR) \cite{wang2004image} and structural similarity index (SSIM) \cite{wang2004image}. For quantitatively evaluating the quality of the generated captions, we use MSCOCO caption evaluation tool\footnote{\url{https://github.com/tylin/coco-caption}}, including Bleu, Meteor, Rouge-L, CIDEr \cite{chen2015microsoft} and Spice \cite{anderson2016spice}.

\paragraph{Competing Methods.} To evaluate the severely image deblurring branch, we compare the proposed S3E-Deblur with the following recent methods on kernel-free blind deblurring: 1) DeblurGAN \cite{kupyn2018deblurgan} (state-of-the-art): A conditional GAN based deblurring model with adversarial and perceptual components. 2) U-Deconv \cite{aittala2018burst}: An U-net \cite{ronneberger2015u} based deconvolution model for both single-image and video deblurrings (we only refer to the former). To evaluate the blurred image captioning branch, we compare the proposed S3E-Deblur with the following representative methods: 1) NIC \cite{vinyals2015show}: A fundamental and mainstream CNN-RNN architecture for image captioning. 2) Toronto \cite{xu2015show}: A spatial attention based captioning model. 3) VP-tree \cite{chen2017structcap}: A visual semantic structure (visual parsing tree) based captioning model. 4) TopDown \cite{anderson2018bottom} (state-of-the-art): A bottom-up and top-down visual attention model at the object/region level. Note that our core contribution is to introduce captioning into deblurring. We are therefore free and open to other cutting-edge image captioning models.

\subsection{Evaluation on Image Deblurring\label{sec:eval_SID}}

The quantitative results on severely blurred image are shown in Tab. \ref{tab:comparison_sota_SID}, where the methods are compared on both LessSev-BlurData and Sev-BlurData. The proposed S3E-Deblur outperforms U-Deconv \cite{aittala2018burst} and DeblurGAN \cite{kupyn2018deblurgan} on both PSNR and SSIM. Especially on Sev-BlurData, S3E-Deblur achieves more highlighted performances, which reflects the apparent superiority of S3E-Deblur on the severely blurred images. Also, the evident promotion can be found between S3E-Deblur and its baseline, DeblurGAN, which manifests the significant role of S3-tree when the semantic contents of the blurry images are ambiguous.  Additionally, Tab. \ref{tab:comparison_sota_SID} also shows that S3E-Deblur co-trained with image captioning branch performs better, which verifies the effectiveness of training image captioning model to improve the semantic understanding performance using S3-tree. The qualitative comparisons are shown in Fig. \ref{fig:visSID_cocoBased}. Clearly, S3E-Deblur can avoid the severe artifacts, especially on the small objects far away, like \emph{wing} in Fig. \ref{fig:visSID_cocoBased} (Top), which verifies the prominent effect of the semantic contents during the model training.

To evaluate the robustness of our model, we further conduct deblurring evaluation on the severely blurred GoPro dataset. The quantitative results are shown in Tab. \ref{tab:comparison_sota_SID_transfer}, where the performances of the proposed S3E-Deblur overpass U-Deconv \cite{aittala2018burst} and DeblurGAN \cite{kupyn2018deblurgan} with large margins. Specially, S3E-Deblur co-trained with the semantic component outperforms the one embedded by pre-trained S3-tree, which manifests the enhanced effect in co-training manner. We present the qualitative results in Fig. \ref{fig:visSID_otherBased}, where the restored qualities of S3E-Deblur are apparently higher than the others. Both the quantitative and the qualitative results reflect both the superiority and robustness of S3E-Deblur, as well as its general applicability.

\subsection{Evaluation on Image Captioning\label{sec:eval_BIC}}

To evaluate the proposed S3-tree model on semantic understanding. Tab. \ref{tab:comparison_sota_BIC} shows the quantitative comparisons among the representative image captioning methods, where the proposed S3E-Deblur (ImC) performs the best in most metrics. Notably, S3E-Deblur (ImC) achieves a larger performance gain when dealing with more severely blurred images. We can conclude that: S3-tree is apparently superior in capturing and representing the semantic contents, and works reasonably well for severely blurred images. To evaluate the performance intuitively, we provide the captioning results in Fig. \ref{fig:VisBIC}. The qualitative comparisons show that the proposed S3E-Deblur (ImC) generates more accurate captions, which further demonstrates the superiority of S3-tree in understanding semantics of the severely blurred images.

\subsection{Ablation Study of S3-tree\label{sec:analysis_SSS-tree}}

We further conduct ablation study of S3-tree model. We visualize the outputs of each node in S3-tree in Fig. \ref{fig:visSSS-tree}. In term of the probability distributions of the semantic items (entities/relations), S3-tree accurately predicts most entity/relation categories (with the top-3 highest probabilities). Additionally, the entities/relations with the second and the third highest probabilities are generally reasonable. It indicates that both the explicit and the implicit semantic contents are well captured. In term of the heatmaps of the feature maps, the activations of nodes cover most of the correct locations, which further verifies the ability of S3-tree on capturing and representing the semantic contents.

\section{Conclusion}

In this paper, we exploited to deblur the severely blurred images with semantic (language) guidance. Our work serves as the first to link the image blurring (ImD) and image captioning (ImC) tasks in a unified framework. To capture and represent the semantic contents in the blurry images, we proposed a novel Structured-Spatial Semantic Embedding model (termed S3E-Deblur), where a novel Structured-Spatial Semantic tree model (S3-tree) bridges ImD and ImC. In particular, S3-tree captures and represents the semantic contents in structured spatial features in ImC, and then embeds the spatial features of the tree nodes into the generator of the GAN based ImD. Co-training on S3-tree, ImC, and ImD is conducted to optimize the overall model in a multi-task end-to-end manner. Extensive experiments on severely blurred MSCOCO and GoPro datasets demonstrate the significant superiority of S3E-Deblur over the state-of-the-arts on both deblurring and captioning tasks.

\newpage
\bibliographystyle{unsrt}
{\small
\bibliographystyle{ieee}
\bibliography{egbib}
}

\end{document}